\documentclass[10pt,twocolumn,letterpaper]{article}

\usepackage{wacv}      % To produce the REVIEW version for the algorithms track
\usepackage{graphicx}
\usepackage{amsmath}
\usepackage{amssymb}
\usepackage{booktabs}
\usepackage{color,soul}
\usepackage{amsfonts}
\usepackage{txfonts}

\usepackage[pagebackref,breaklinks,colorlinks]{hyperref}

\usepackage[table,xcdraw]{xcolor}   % For coloring rows

\usepackage[capitalize]{cleveref}
\crefname{section}{Sec.}{Secs.}
\Crefname{section}{Section}{Sections}
\Crefname{table}{Table}{Tables}
\crefname{table}{Tab.}{Tabs.}

\def\wacvPaperID{1} % *** Enter the WACV Paper ID here
\def\confName{WACV}
\def\confYear{2024}

\begin{document}

\title{An Exploratory Study on Human-Centric Video Anomaly Detection through Variational Autoencoders and Trajectory Prediction}

\author{Ghazal Alinezhad Noghre
\and
Armin Danesh Pazho
\and
Hamed Tabkhi
\and
University of North Carolina at Charlotte\\
Charlotte, NC, USA\\
{\tt\small \{galinezh, adaneshp, htabkhiv\}@uncc.edu}
% {\tt\small adaneshp@uncc.edu}
% {\tt\small htabkhiv@uncc.edu}
}
\maketitle

% University of North Carolina at Charlotte\\
% Charlotte, NC, USA\\
% {\tt\small galinezh@uncc.edu}
% University of North Carolina at Charlotte\\
% Charlotte, NC, USA\\
% {\tt\small adaneshp@uncc.edu}

% University of North Carolina at Charlotte\\
% Charlotte, NC, USA\\
% {\tt\small htabkhiv@uncc.edu}

%%%%%%%%% ABSTRACT
\begin{abstract}

Video Anomaly Detection (VAD) represents a challenging and prominent research task within computer vision. In recent years, Pose-based Video Anomaly Detection (PAD) has drawn considerable attention from the research community due to several inherent advantages over pixel-based approaches despite the occasional suboptimal performance. Specifically, PAD is characterized by reduced computational complexity, intrinsic privacy preservation, and the mitigation of concerns related to discrimination and bias against specific demographic groups. This paper introduces TSGAD, a novel human-centric Two-Stream Graph-Improved Anomaly Detection leveraging Variational Autoencoders (VAEs) and trajectory prediction. TSGAD aims to explore the possibility of utilizing VAEs as a new approach for pose-based human-centric VAD alongside the benefits of trajectory prediction. We demonstrate TSGAD's effectiveness through comprehensive experimentation on benchmark datasets. TSGAD demonstrates comparable results with state-of-the-art methods showcasing the potential of adopting variational autoencoders. This suggests a promising direction for future research endeavors. The code base for this work is available at \url{https://github.com/TeCSAR-UNCC/TSGAD}.

\end{abstract}

%%%%%%%%% BODY TEXT
\section{Introduction}
\label{sec:intro}
In recent years, surveillance cameras have been proliferation; nevertheless, the available human resources are insufficient for real-time monitoring and expeditious, judicious response to the voluminous video feed generated by these cameras \cite{pazho2023ancilia}. Furthermore, there may exist an inherent bias in decisions made by humans. Hence, as Artificial Intelligence (AI) continues to advance, the integration of smart technologies for the detection of anomalous behaviors has garnered significant attention across various communities.

Anomaly detection can refer to a wide range of applications \cite{hojjati2023multivariate, pazho2023survey,caetano2022deep,ho2023self, sliti2023f}. One of the main subsets is the domain of human-centric video anomaly detection that has been examined from two primary perspectives: pose-based video anomaly detection \cite{yu2023regularity, markovitz2020graph} and pixel-based video anomaly detection \cite{reiss2022attribute, wang2022video}. While pixel-based approaches typically demonstrate superior detection accuracy, Pose-based Anomaly Detection (PAD) has attracted considerable research interest due to reduced computational complexity, inherent privacy preservation, and robustness to visual variations and background noise \cite{liu2020privacy, cormier2022we, buet2022towards}. However, PAD methods may suffer from limited information and reliance on accurate pose estimation. The choice between these approaches should consider the specific application's requirements, balancing the need for privacy, efficiency, and the nature of targeted anomalies. In this article, our focus is pose-based approaches.

Video anomaly detection presents an intrinsic challenge as it inherently constitutes an open-set problem characterized by the potential emergence of diverse normal and abnormal behaviors within the real-world setup, driven by the complexity of human behavior. The supervised training, often reliant on data that may inadequately represent the entirety of anomalies, suffers from limited generalizability \cite{noghre2023understanding}. In response to this challenge, the field employs unsupervised learning techniques as the most common approach \cite{hojjati2022self} to improve the efficacy of anomaly detection models. We embrace the unsupervised paradigm to mitigate this challenge in line with prior scholarly investigations.

% Various strategies have been employed to address the challenge of PAD, encompassing techniques such as reconstruction, prediction, and distribution analysis. This study introduces TSGAD, a novel approach that amalgamates reconstruction, distribution analysis, and trajectory prediction, offering promising avenues for improving PAD performance. TSGAD represents a departure from conventional methodologies by incorporating Variational Autoencoders (VAEs) \cite{kingma2013auto} as a pivotal constituent for PAD. On top of that, we propose utilizing a State-of-the-Art (SotA) trajectory prediction model \cite{alinezhad2023pishgu} as a complementary branch for anomaly detection. The derivative of the human pose concerning the temporal axis yields the person's trajectory, precisely defined as the temporal evolution of the spatial displacement of the central keypoint (middle of the hips). Adopting trajectory prediction for PAD is motivated by two main reasons. Firstly, in instances involving individuals positioned at a significant distance from the camera, trajectory data exhibits reduced noise levels compared to pose data. Secondly, the trajectory branch can capture social interactions missing from most pose-based methods through Graph Isomorphism.

This study introduces TSGAD, a novel approach that amalgamates reconstruction, distribution analysis, and trajectory prediction, offering promising avenues to enhance PAD performance. TSGAD departs from conventional methodologies by incorporating Variational Autoencoders (VAEs) \cite{kingma2013auto}. Additionally, we propose the use of a State-of-the-Art (SotA) trajectory prediction model \cite{alinezhad2023pishgu} as a complementary branch for PAD. The trajectory, derived from the human pose, is defined as the temporal evolution of the spatial displacement of the central keypoint (middle of the hips). Motivations for using trajectory are two-fold; firstly, in instances involving individuals positioned at a significant distance from the camera, trajectory data exhibits reduced noise levels compared to pose. Secondly, the chosen trajectory prediction model can capture social interactions missing from most pose-based methods through Graph Isomorphism.

We evaluate the proposed TSGAD method through comprehensive experiments. To ensure a thorough analysis, unlike most previous works, we employ not only the conventional Area Under the Receiver Operating Characteristic Curve (AUC-ROC) metric but also supplementary metrics, including Area Under the Precision-Recall Curve (AUC-PR) and Equal Error Rate (EER). These additional metrics provide a more nuanced understanding of our model's performance, addressing aspects that AUC-ROC may not fully represent. TSGAD attains AUC-ROC values of $80.67\%$, $81.77\%$, and $69.55\%$ on three well-known anomaly detection benchmarks namely ShanghaiTech \cite{liu2018future}, Human Related ShanghaiTech \cite{morais2019learning}, and CHAD \cite{danesh2023chad} respectively. This exploratory study and its results showcase our method's ability to compete with SotA models in the field of anomaly detection, affirming its potential as a novel avenue for future endeavors and enhancement in PAD.

The contributions of this paper are as follows:
\begin{itemize}
    \item Investigating the fusion of prediction-based and reconstruction-based approaches utilizing Variational Autoencoders (VAE) for human-related anomaly detection.
    \item Exploring the benefits of using social interaction-aware trajectory prediction for anomaly detection and proposing an integrated approach that combines pose and trajectory methods for comprehensive anomaly detection.
    \item Conducting a thorough evaluation of the proposed solutions using a comprehensive range of metrics to gain deeper insights into the merits and limitations of the design under consideration.
    \item Empirical analysis of different pose anomaly score formulations to assess their impact on anomaly detection performance.
\end{itemize}

\section{Related Works}
\label{sec:related}
\subsection{Pixel-based Approaches}
\cite{reiss2022attribute} proposes a multi-branch design for anomaly detection. The proposed method is based on the idea that anomalies can be detected using abrupt changes in velocity, pose, and deep features extracted from input frames. \cite{wang2022video} tackles anomaly detection by solving a spatio-temporal jigsaw puzzle. The jigsaw solver is only trained using normal videos. The permutation predictions from the solver are used as a measure for anomaly detection. \cite{cao2022context} introduced a two-stream framework for anomaly detection. The context recovery stream predicts the future video frames and the knowledge recovery stream compares the video snippet to the knowledge gathered from training on normal videos. \cite{georgescu2021anomaly} also proposes a multi-branch multi-task design. The first branch learns to predict the arrow of time, assuming that detecting the arrow of time is harder for anomalous behaviors. The second branch tries to detect irregular motion in the input sequence. The final branch reconstructs the detected objects bounding boxes. The model is trained on normal samples, and in the inference stage, the score from all these tree branches is fused to form the final score.

\subsection{Pose-based Approaches}
Normal Graph \cite{luo2021normal} leverages spatial-temporal graph convolutional network for predicting future pose segments trained on the normal data. The predicted pose segments are then compared to the ground truth, and the Mean Squared Error (MSE) loss is used as the anomaly score. Since the model is only trained on normal data, it cannot predict anomalous movements accurately. Hence the drastic difference between the prediction and actual future movement can reveal anomalous behavior.\cite{rodrigues2020multi} uses a similar prediction-based approach. \cite{rodrigues2020multi} not only predicts future pose sequences but also leverages a past prediction module for multi-scale past/future prediction to enhance the accuracy of the final anomaly detection model. \cite{zeng2021hierarchical} similarly proposes a prediction-based method but with three branches for predicting future pose, trajectory, and the motion vector. MPED-RNN \cite{morais2019learning} proposes a Gated Recurrent Unit-based (GRU) encoder-decoder structure with two decoder heads: a predicting and a reconstructing head. The reconstruction and prediction scores are then fused to generate the anomaly score. Similar to MPED-RNN, \cite{li2022human} uses both reconstruction and prediction for anomaly detection, but \cite{li2022human} uses Long Short-Term Memory (LSTM) units instead of GRUs. MemWGAN-GP\cite{li2023human} also uses a similar dual-head decoder structure upgraded with a modified version of the Wasserstein generative adversarial network \cite{pmlr-v70-arjovsky17a} for improving the prediction and reconstruction quality. GEPC \cite{markovitz2020graph} uses a spatio-temporal graph autoencoder combined with a clustering layer for assigning soft probabilities to the input pose segments. The output probabilities are a measure of anomalous behavior.

\section{Preliminaries}
\label{sec:preliminaries}
\subsection{Variational Autoencoders}
Variational Autoencoder (VAE) first introduced by \cite{kingma2013auto} is a type of generative model initially designed for probabilistically generating content. They have gained popularity for their capability to model complex data distribution in an unsupervised manner. Similar to Autoencoders (AEs), VAEs can be used for feature learning, dimensionality reduction, denoising, etc. However, VAEs can be more advantageous for anomaly detection due to their probabilistic nature and ability to capture intricate data distributions. In VAEs, the objective function consists of two terms: a reconstruction term similar to AEs and a regularization term as depicted by \cref{eq:loss}:

\begin{equation}
\label{eq:loss}
L_{\mathrm{VAE}}=L_{\text {reconstruction }}+L_{\text {regularization }}
\end{equation}

$L_{VAE}$ indicates Evidence Lower Bound (ELBO) which is maximized during the training. The reconstruction term is defined as the log-likelihood of the observed data given the latent variable:

\begin{equation}
\label{eq:loss_recon}
L_{reconstruction}=\frac{1}{N} \sum_{n=1}^N\mathbb{E}_q\left[\log p\left(x_n \mid z\right)]\right.
\end{equation}

where $N$ is the number of data samples, $x_n$ is the data sample and $z$ is the latent variable. On the other hand, the regularization term encourages the latent space to have a specific structure, typically a multivariate Gaussian distribution. This objective is achieved using  Kullback-Leibler (KL) divergence between the approximate posterior distribution $q(z \mid x)$ and the prior distribution $p(z)$:

\begin{equation}
\label{eq:loss_kl}
L_{regularization}=-\frac{1}{N} \sum_{n=1}^N \operatorname{KL}\left(q\left(z \mid x_n\right) \| p(z)\right)
\end{equation}

Another variant of VAEs named $\beta$-VAE \cite{burgess2018understanding} was later introduced to learn a more disentangled representation in the latent space. This goal is achieved by highlighting the regularization term in \cref{eq:loss} by adding a multiplier $\beta$ with setting $\beta$ values greater than $1$. Later, TC-VAE \cite{chen2018isolating} introduces a new formulation for achieving better disentanglement:

\begin{equation}
\label{eq:tc_loss}
\begin{split}
L_{TC-VAE}:=\mathbb{E}_{q(z \mid n) p(n)}[\log p(n \mid z)]-\alpha I_q(z ; n) \\
-\beta \operatorname{KL}\left(q(z) \| \prod_j q\left(z_j\right)\right)-\gamma \sum_j \operatorname{KL}\left(q\left(z_j\right) \| p\left(z_j\right)\right)
\end{split}
\end{equation}

Where $n$ is a distinct integer index assigned to every training datapoint establishing a random variable that uniformly covers the range from $1$ to $N$. Keep in mind that $q(z,n) = q(z|n)p(n)=q(z|n)\frac{1}{N}$. The first term corresponds to the reconstruction loss. The second term is the Index-Code Mutual Information, which is the mutual information between the data variable and latent variable or $I_q(z;n)$. It is shown that maximizing this term results in learning more disentangled latent representations based on previous studies \cite{chen2016infogan, burgess2018understanding}. Total correlation calculated by the third term in \cref{eq:tc_loss} quantifies the dependence between variables. Optimizing this term leads to learning independent factors in the data distribution. The final term denoted as dimension-wise KL, is responsible for maintaining congruence between the latent dimensions and their respective prior distributions. $\alpha$, $\beta$, and $\gamma$ are adjustable multipliers chosen by the requirements.

\subsection{Trajectory Prediction}
A trajectory prediction problem is forecasting a subject's future position based on the observed trajectory. Trajectory prediction has many real-world computer vision applications such as pedestrian safety, transportation safety, intelligent traffic
monitoring, and video surveillance \cite{georgiou2018moving, wang2019exploring}. Predicted trajectory can also be used for anomaly detection; sudden changes in the trajectory can be an indication of anomalous events. 

We leverage Pishgu \cite{alinezhad2023pishgu} for the purpose of anomaly detection. Pishgu uses the Graph Isomorphism Network (GIN) to capture the interdependencies between subjects available in the scene and constructs latent representations considering both social interactions and the movement history of the subject. In the next step, an attentive Convolutional Neural Network (CNN) is used for capturing temporal relations and constructing the final predicted trajectories. 

\section{TSGAD}
\label{sec:methodology}
\begin{figure*}[ht!]
    \centering
    % \resizebox{1\linewidth}{!}{
    \includegraphics[clip,trim={20 18 26 18},width=1\textwidth]{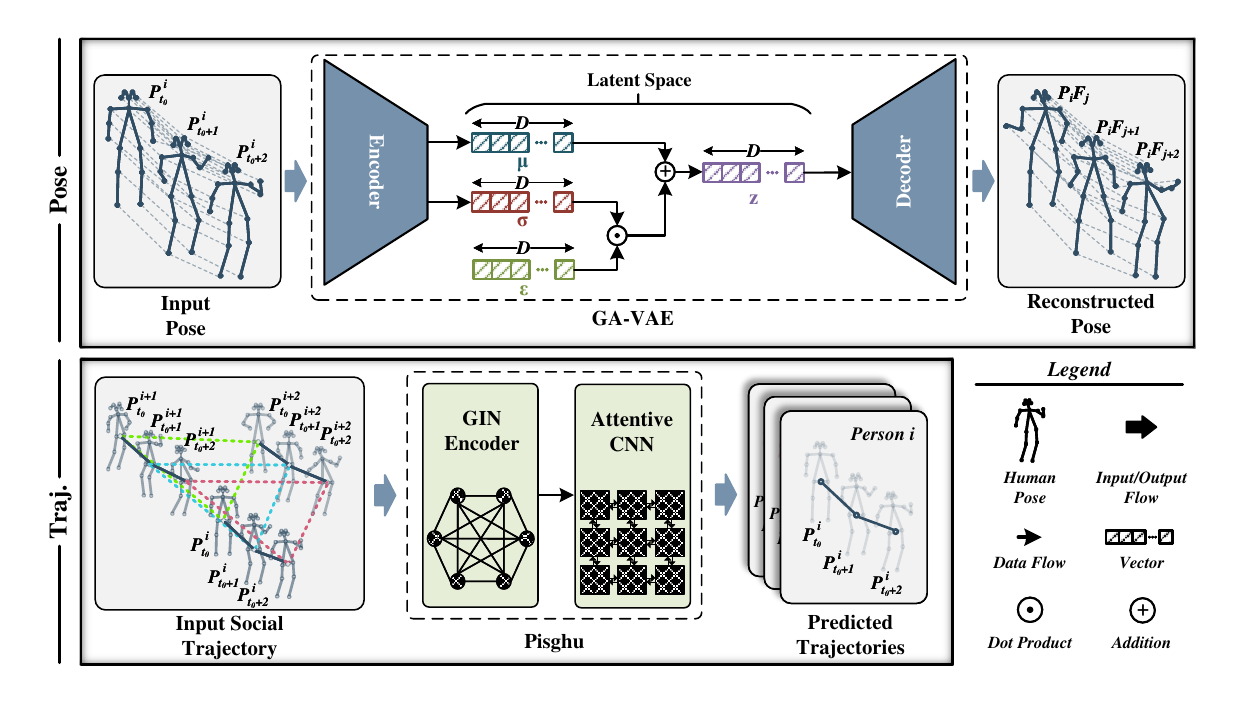}
    % }
    \caption{TSGAD architecture. The upper branch utilizes Graph Attentive Variational Autoencoder (GA-VAE) for learning the characteristics of normal human behavior distribution in an unsupervised manner. The lower branch leverages a SotA trajectory prediction method, namely Pishgu \cite{alinezhad2023pishgu}, for learning how to predict normal trajectories. $P_t^i$ denotes the $i^{th}$ person at time $t$, and $D$, $\mu$, and $\sigma$ refer to the latent representation's dimensions, mean, and variance. $z$ follows a normal distribution with $z \sim (0, I)$, where $I$ is the identity matrix.}
    \label{fig:TSGAD}
\end{figure*}

In this section, we introduce our proposed TSGAD methodology.
\subsection{Problem Formulation}
As depicted in \cref{fig:TSGAD}, TSGAD has two branches. The top branch uses a window of size $T_{in}$ of observed poses as input:

\begin{equation}
  \mathcal{P}^i = [P_{t_0}^i, P_{t_0 +1}^i, ..., P_{t_0 +T_{in}-1}^i]  
\end{equation}
where $P_{t_0}^i$ shows the pose of person $i$ in frame ${t_0}$. A sequence of the position of the center of a person is used as the input of the trajectory branch:

\begin{equation}
  \mathcal{C}^i = [C_{t_0}^i, C_{t_0+1}^i, ..., C_{t_0+T'_{in}-1}^i]  
\end{equation}

where $C_{t_0}$ is the location of the center of person $i$ in frame $t_0$ and $T'_{in}$ is the input window size for the trajectory prediction branch. Pose branch and trajectory branch output $S_{Pose}^i$ and $S_{Traj}^i$ respectively associated with the $i^{th}$ person. The final output of the model is constructed by combining normalized $S_{Pose}^i$ and $S_{Traj}^i$:

\begin{equation}
    \label{final_score}
    S_{Final}^i = a \cdot Norm(S_{Pose}^i) + b \cdot Norm(S_{Traj}^i)
\end{equation}

where $a$ and $b$ are multipliers calculated by dividing the AUC-ROC of each branch by the sum of the AUC-ROC of both branches. In the final step, the maximum score over all available subjects in the scene is calculated and considered as the anomaly score associated with a frame:
\begin{equation}
     S_{Final} = max_{i \in N}( S_{Final}^i )
\end{equation}

\subsection {Archietcture}
As depicted in \cref{fig:TSGAD}, the proposed model consists of two branches; the top branch uses pose data and Graph Attentive Variational Autoencoder (GA-VAE) to capture the distribution of normal behavior. The bottom branch uses the state-of-the-art trajectory prediction method Pishgu \cite{alinezhad2023pishgu} to predict future trajectories. Deviation from predicted trajectories is used as a measure of anomaly detection. 
\subsubsection{GA-VAE}
\begin{figure}[]
    
    \centering
    % \resizebox{1\linewidth}{!}{
    \includegraphics[clip,trim={23 20 25 18},width=1\columnwidth]{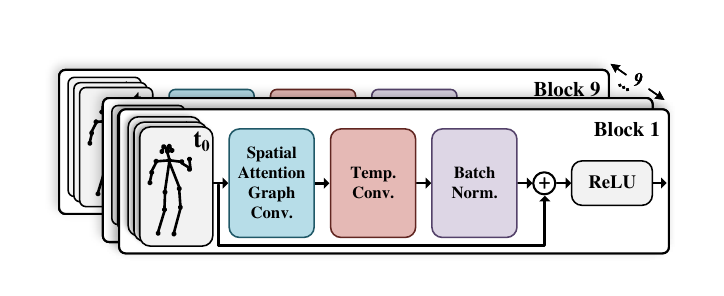}
    % }
    \caption{Nine layers of spatio-temporal graph convolution blocks are stacked forming the GA-VAE encoder. Each block consists of a spatial attention graph convolution followed by temporal convolution, batch normalization, a residual connection, and a final activation function. }
    \label{fig:STGCN}
\end{figure}

In order to capture the relationships between joints in human pose, we choose to represent the human pose using a spatio-temporal graph formulation. The joints are considered the nodes of the graph, and the edges represent the physical limbs and learned motion dependencies necessary for modeling the human pose effectively. In the context of video anomaly detection, it is imperative to incorporate temporal edges to represent the temporal interdependencies among frames. Thus, the resulting graph is a spatio-temporal graph that formulates human motion. We propose building a deep variational autoencoder leveraging Spatial-Temporal Graph Convolution (ST-GCN) blocks \cite{yu2017spatio}. \cite{markovitz2020graph} extended ST-GCN blocks by adding more sophisticated spatial attention, including three GCN blocks for better capturing physical relations, dataset-level keypoint relations, and sample-specific relations. We chose a symmetric design for the VAE with both Graph Attentive Probabilistic encoder (GA-VAE encoder) and Graph Attentive Probabilistic decoder (GA-VAE decoder) having $9$ layers of modified ST-GCN blocks, demonstrated in \cref{fig:STGCN}. Unlike previous works that use ST-GCN for processing input pose data, we adopt a probabilistic approach using ST-GCN blocks for constructing a VAE. We consider the prior distribution to be a normal distribution to match real-world human behavior. The probabilistic design is instrumental in capturing the inherent distribution of input data, thereby enhancing the modeling of normal behavior and consequently leading to improved performance in the context of anomaly detection. 

The training procedure is conducted in an unsupervised fashion, wherein the GA-VAE is trained on the training set, which exclusively includes normal behavior exemplars. During the GA-VAE training phase, we implement the Evidence Lower Bound (ELBO) loss as introduced in \cref{eq:tc_loss} with the same setup for multipliers as \cite{chen2018isolating}. Throughout the training, the model minimizes the negative $L_{GA-VAE}$ to encourage the understanding of normal behavior patterns. After the model is trained, for each datapoint in the training set, the corresponding normal distribution parameters ($\mu_n$ and $\sigma_n$) are concatenated and averaged over the training set to find an Aggregated Parameter Index (API) that is a single vector representing the characteristics of normal behavior:

\begin{equation}
\label{eq:api}
    API = \frac{1}{N} \sum_{n \in N} (\mu_n \mathbin\Vert \sigma_n)
\end{equation}

where N is the number of datapoints in the training set. 

\cref{fig:inference} shows the inference process. Each datapoint in the test set is passed through the trained GA-VAE encoder to map to a normal distribution parameterized by $\mu_n$ and $\sigma_n$. To calculate the anomaly score, we measure the deviation from API:

\begin{equation}
    S_{Pose} =\sqrt{\sum_{j=1}\left((\mu_n \mathbin\Vert \sigma_n)_j-API_j\right)^2}
\end{equation}

where $(\mu_n \mathbin\Vert \sigma_n)_j$ is the $j^{th}$ dimension of the latent parameters of input datapoint. 
\subsubsection{Trajectory Prediction for Anomaly Detection}

As illustrated in \cref{fig:TSGAD}, we advocate the incorporation of a trajectory prediction model within the context of anomaly detection. The primary goal is to introduce the dynamics of interactions between subjects, thereby extracting valuable insights for anomaly detection. The trajectory prediction branch, fundamentally concerned with modeling the collective movements of individuals within the scene, complements the pose-based anomaly detection approach. This approach is introduced to mitigate challenges associated with pose estimation inaccuracies. Consequently, the trajectory perspective provides a complimentary and holistic representation of the scene, contributing to improved anomaly detection capabilities.

We adopt the state-of-the-art trajectory prediction model Pishgu, as introduced by \cite{alinezhad2023pishgu} for the specific application of anomaly detection. We train Pishgu exclusively on the training set, comprising instances of normal behavior, with a focus on capturing the nuanced features of typical movements. The training process involves the optimization of the Mean Squared Error (MSE) loss function.

\begin{equation}
\mathrm{MSE}=\frac{1}{N} \sum_{n=1}^N\left(Y_n-\hat{Y}_n\right)^2
\end{equation}

where $\hat{Y}$ is the predicted trajectory and $Y$ is the actual coordinates of a person. 

\begin{figure}[]
    
    \centering
    % \resizebox{1\linewidth}{!}{
    \includegraphics[clip,trim={18 25 18 17},width=1\columnwidth]{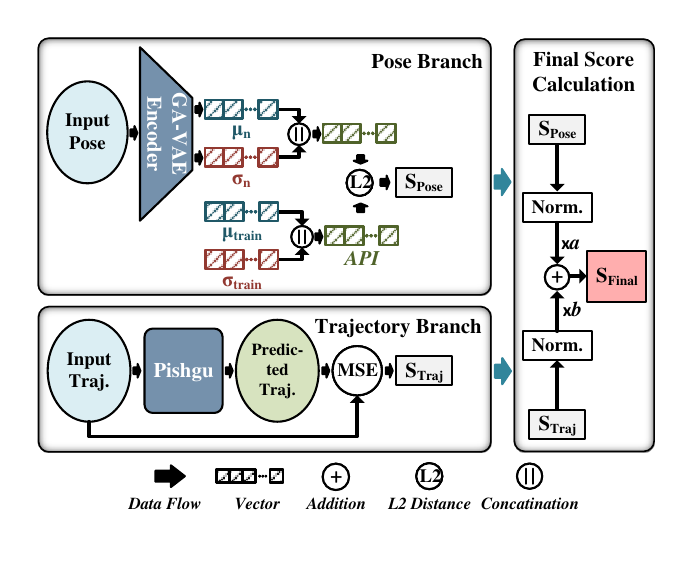}
    % }
    \caption{The inference phase. The deviation from API in the latent space is used for calculating the pose score ($S_{Pose}$). The difference between the predicted trajectory and the actual trajectory measured by MSE is used to form a trajectory score ($S_{Traj}$). The weighted sum of these normalized scores forms the final anomaly score. $\mu_n$, $\sigma_n$, and API refer to the mean, and variance of the latent representation and Aggregated Parameter Index defined in \cref{eq:api} respectively.}
    \label{fig:inference}
\end{figure}

In the inference phase, as illustrated by \cref{fig:inference}, the predicted trajectories for each datapoint are compared to the corresponding actual trajectories. The deviation is measured using MSE loss and used as an anomaly score $S_{Traj}$. 

\section{Experimental Setup}
\label{sec:setup}
In this section, we focus on different aspects of our experimental setup. All the trainings and evaluations have been conducted on a computational workstation featuring three NVIDIA RTX A6000 GPUs and an AMD EPYC 7513 32-core CPU with 256 gigabytes of memory.

\subsection{Datasets}

\subsubsection{ShanghaiTech Campus (SHT)}

The ShanghaiTech Campus (SHT) dataset \cite{liu2018future} serves as the principal benchmark for video anomaly detection. This dataset comprises more than 317,000 frames spanning 13 distinctive scenes. SHT is partitioned into an unsupervised subset, including in excess of 274,000 normal training frames and 42,883 normal and anomalous frames for testing purposes. For the purposes of our investigation, we adopt the unsupervised split to facilitate a meaningful comparison with prior research endeavors. PAD methodologies assess their models using this dataset due to its expansive scale and the presence of videos with sufficient quality recorded at 24 frames per second (FPS) for pose extraction. Additionally, the types of anomalies in this dataset are also a good representative of real-world scenarios. Accordingly, we integrate the SHT dataset into our experimental framework, employing a methodology consistent with previous SotA studies \cite{yu2023regularity} for pose extraction and tracking \cite{li2019crowdpose}, thus ensuring equitable comparisons.

\subsubsection{HR-ShanghaiTech (HR-SHT)}

The dataset introduced in \cite{morais2019learning} represents a specialized adaptation of the ShanghaiTech Campus (SHT) dataset, specifically tailored for human-related anomaly detection. It is essential to underscore that the sole differentiation between this dataset and the original SHT dataset lies in its exclusive concentration on anomalies related to human activities.

\subsubsection{Charlotte Anomaly Dataset (CHAD)}

Pazho et.al. \cite{danesh2023chad} introduce the Charlotte Anomaly Dataset (CHAD), a high-resolution multi-camera dataset for video anomaly detection in real-world scenarios. It comprises approximately 1.15 million frames, encompassing 1.09 million normal frames and 59,000 anomalous frames with detailed annotations for human detection, tracking, and pose. It is suitable for both unsupervised and skeleton-based anomaly detection methods and emphasizes the use of multiple metrics for benchmarking, discussed in \cref{sec:metrics}. CHAD simulates a real parking lot surveillance environment with four high-resolution cameras recording at 30 FPS, and a diverse set of actors engaging in normal and anomalous behaviors across 22 different anomaly classes. It stands out as the largest anomaly detection dataset with pose and tracking annotations. We exclusively employ the official unsupervised split of CHAD. We choose to conduct experiments on this dataset due to its role in establishing a standardized benchmark for pose-based anomaly detection, thereby mitigating variations due to the quality of extracted poses by different methods.

\subsection{Training Setup}

To optimize hyperparameters, a grid search methodology has been employed, systematically exploring parameter combinations. All trainings have been conducted 5 times to ensure stability. Adam optimizer is used in all training setups. 

\subsubsection{SHT and HR-SHT}

For SHT and HR-SHT datasets, a batch size of 256 is employed for processing the data, while a dropout rate of 0.3 is applied to regularize the neural network. The model undergoes training for 20 epochs with an initial learning rate of 0.005 and a decay factor of 0.99. Also, a weight decay of 5.0e-05 to prevent overfitting. The input window size for set to 24 frames, allowing the model to capture meaningful patterns within the data over a span of 1 second.

For training the trajectory branch, we follow the setup of \cite{alinezhad2023pishgu}; we used the input window size of 16 frames and the output prediction horizon of 14 frames and trained the model for 80 epochs with a learning rate of 0.02 and batch size of 64. 

\subsubsection{CHAD}
For the CHAD dataset, a similar configuration is employed. A batch size of 256 is used, and a dropout rate of 0.3 is applied for regularization. The training also spans 20 epochs, with a slightly higher initial learning rate of 0.01, which still decays by a factor of 0.99. Similar to the first setup a weight decay of 5.0e-05 is employed for regularization. In this case, the input window size is set to 1 second or 30 frames.

When training the trajectory branch on CHAD, we use similar parameters as training on SHT with a window size set to 16 for input data and a prediction horizon of 14 for the output. The training process spanned 80 epochs, employing a learning rate of 0.02, a batch size of 64, and the Adam optimizer.

\subsection{Metrics}
\label{sec:metrics}

\subsubsection{AUC-ROC}
The Area Under the Receiver Operating Characteristic (AUC-ROC) curve is used to evaluate the accuracy of binary classification models. The ROC curve represents the trade-off between a model's true positive rate (sensitivity) and its false positive rate (1-specificity) at various classification thresholds. A higher AUC-ROC score indicates a better model performance.

\subsection{AUC-PR}
The Area Under the Precision-Recall Curve (AUC-PR) quantifies the quality of binary classification, particularly in imbalanced datasets. AUC-PR measures the area under the precision-recall curve, where precision represents the ratio of true positive predictions to the total positive predictions, and recall (sensitivity) quantifies the model's ability to capture all true positive instances. A higher AUC-PR value indicates a better-performing model, as it reflects a higher precision-recall trade-off, signifying superior discrimination and a more effective approach for problems where false positives can be costly or misleading.

\subsection{EER}
The Equal Error Rate (EER) quantifies the point at which the False Acceptance Rate (FAR) and False Rejection Rate (FRR) are equal for a binary classification model. EER identifies the threshold at which the decision boundary balances the rate of false positives vs. false negatives. Please note that EER alone is insufficient \cite{sultani2018real} but when combined with other metrics, offers valuable insights. A lower EER signifies improved system accuracy, as it indicates a better equilibrium between security and usability.

\section{Results}
\label{sec:results}
\subsection{Comparison with State-of-the-Art Approaches}

\begin{table}[]
\centering
\caption{AUC-ROC compared on SHT \cite{liu2018future}, HR-SHT \cite{morais2019learning}, and CHAD \cite{danesh2023chad} datasets.}
\label{tab:roc}
% \resizebox{\columnwidth}{!}{%
\begin{tabular}{lccc}
\toprule[\heavyrulewidth] \midrule
\textbf{Methods}     & \textbf{SHT} & \textbf{HR-SHT} & \textbf{CHAD} \\ \midrule
\textbf{MPED-RNN} \cite{morais2019learning}    & 73.40        & 75.40           & -             \\
\textbf{GEPC} \cite{markovitz2020graph}        & 75.50        & -               & 64.90          \\
\textbf{PoseCVAE} \cite{jain2021posecvae}   & 74.90        & 75.70           & -             \\
\textbf{MSTA-GCN} \cite{chen2023multiscale}    & 75.90        & -               & -             \\
\textbf{MTP} \cite{rodrigues2020multi}        & 76.03        & 77.04           & -             \\
\textbf{HSTGCNN} \cite{zeng2021hierarchical}    & {81.80}        & {83.40}           & -  \\          
\textbf{STGformer} \cite{huang2022hierarchical}   & 82.90        & {86.97}           & -             \\ \midrule
%\textbf{MoPRL}       & 83.35        & 84.4            & -             \\ \midrule
\rowcolor[HTML]{EFEFEF}
\textbf{TSGAD-Pose (Ours)} & {80.59}        &    81.52        & {59.30}         \\  
\rowcolor[HTML]{EFEFEF}
\textbf{TSGAD-Traj (Ours)} & 67.78        &      68.45      & {69.55}         \\
\rowcolor[HTML]{EFEFEF}
\textbf{TSGAD (Ours)} & {80.67}        & {81.77}         & {66.49}         \\ \midrule \bottomrule[\heavyrulewidth]
\end{tabular}%
% }
\end{table}

The AUC-ROC metric stands as the most extensively investigated performance measure within the realm of PAD. \cref{tab:roc} presents the investigation across diverse datasets revealing promising results for TSGAD, approaching SotA performance. This underscores the potential for further exploration of the innovative approach that integrates VAEs and probabilistic modeling, along with the fusion of pose and trajectories.

As illustrated in Table \ref{tab:roc}, it is evident that distinct branches exert varying influences, and the outcomes manifest their complementary nature. For instance, in the context of the SHT dataset \cite{liu2018future}, the pose and trajectory branches individually attain AUC-ROC values of $80.59\%$ and $67.78\%$, respectively. However, the combination of these branches yields the most favorable results, implying that each branch addresses a complementary subset of anomalies, and their combined operation aims at mutual enhancement. The same trend can be seen for HR-SHT \cite{morais2019learning} as well.

Conversely, when examining the results for CHAD \cite{danesh2023chad}, the AUC-ROC values distinctly indicate that a trajectory-based approach is notably more well-suited for this particular environment. This observation may serve as an indicator of a potentially noisier environment or less precise pose annotations. As previously discussed, the inherent robustness of trajectory data, when contrasted with pose information, likely contributes to the superior results achieved with trajectory-based anomaly detection in this scenario. The incorporation of both pose and trajectory components appears to diminish the overall performance. Consequently, the selection of the most appropriate model depends on the intrinsic characteristics of the given environment. It is imperative to undertake comprehensive explorations involving diverse branches to tailor the model effectively to the specific requirements of the environment under consideration.

\subsection{Detailed Analysis of Supplementary Metrics}

\begin{table*}[]
\centering
% \caption{AUC-PR and EER of our design compared to GEPC \cite{markovitz2020graph}. The numbers reported for GEPC are based on \cite{danesh2023chad}.}
\caption{AUC-PR and EER of our design compared to GEPC \cite{markovitz2020graph} on SHT \cite{liu2018future}, HR-SHT \cite{morais2019learning}, and CHAD \cite{danesh2023chad} datasets.}
\label{tab:all_metrics}
% \resizebox{\linewidth}{!}{%
\begin{tabular}{@{}ccccccc@{}}
\toprule[\heavyrulewidth] \midrule
            & \multicolumn{2}{c}{\textbf{SHT}} & \multicolumn{2}{c}{\textbf{HR-SHT}} & \multicolumn{2}{c}{\textbf{CHAD}} \\ \midrule
            & \textbf{AUC-PR \textuparrow}  & \textbf{EER \textdownarrow}  & \textbf{AUC-PR \textuparrow}   & \textbf{EER \textdownarrow}    &  \textbf{AUC-PR \textuparrow}  & \textbf{EER \textdownarrow}  \\ \midrule
\textbf{GEPC} \cite{markovitz2020graph}       & 65.70   & 0.31   & -    & - & {58.70}   & 0.38    \\ \midrule
\rowcolor[HTML]{EFEFEF}
\textbf{TSGAD-Pose (Ours)} & 72.20   & 0.25   & 72.07    & 0.25     & 53.69   & 0.41    \\
\rowcolor[HTML]{EFEFEF}
\textbf{TSGAD-Traj (Ours)} & 61.26   & 0.38   &   61.32  &  0.38   & 66.97   & 0.36   \\
\rowcolor[HTML]{EFEFEF}
\textbf{TSGAD (Ours)} & 73.86   & 0.25  & 74.20    & 0.25   &{62.18}   & {0.38}    \\ \midrule \bottomrule
\end{tabular}%
% }
\end{table*}

\begin{table*}[]
\footnotesize
\centering
% \caption{AUC-PR and EER of our design compared to GEPC \cite{markovitz2020graph}. The numbers reported for GEPC are based on \cite{danesh2023chad}.}
\caption{Comparing different methods of calculating pose anomaly score ($S_{Pose}$) on SHT \cite{liu2018future}, HR-SHT \cite{morais2019learning}, and CHAD \cite{danesh2023chad} datasets.}
\label{tab:ablation}
% \resizebox{\linewidth}{!}{%
\begin{tabular}{@{}cccccccccc@{}}
\toprule[\heavyrulewidth] \midrule
            & \multicolumn{3}{c}{\textbf{SHT}} & \multicolumn{3}{c}{\textbf{HR-SHT}} & \multicolumn{3}{c}{\textbf{CHAD}} \\ \midrule
            & \textbf{AUC-ROC \textuparrow} & \textbf{AUC-PR \textuparrow}  & \textbf{EER \textdownarrow}  & \textbf{AUC-ROC \textuparrow} &  \textbf{AUC-PR \textuparrow}   & \textbf{EER \textdownarrow}   & \textbf{AUC-ROC \textuparrow} &  \textbf{AUC-PR \textuparrow}  & \textbf{EER \textdownarrow}  \\ \midrule

\textbf{GA-VAE } &  80.59 & 72.20   & 0.25  & 81.52 & 72.07    & 0.25    & 59.30 & 53.69   & 0.41    \\
\textbf{GA-VAE-ELBO } & 76.08 & 69.14   & 0.30   & 76.91 & 69.33  &  0.30  & 59.77  &   54.19 & 0.43     \\ \midrule \bottomrule
\end{tabular}%
% }
\end{table*}

While the AUC-ROC metric provides valuable insights into the effectiveness of binary classifiers, its applicability diminishes when confronted with imbalanced datasets \cite{he2013imbalanced}. Conversely, AUC-PR exhibits greater resilience in the presence of imbalanced data, thereby aiding in a deeper comprehension of the model's underlying characteristics. In addition to examining AUC-PR, we report EER to not only gain better insights into the sensitivity-specificity balance but also assess the real-world practicality of our model. AUC-PR and EER are only compared to GEPC \cite{markovitz2020graph} as it is the only model that its performance was reported for these metrics in \cite{danesh2023chad}.

As evident from the comprehensive metrics presented in \cref{tab:all_metrics}, the behavior of the AUC-PR aligns with the trends observed in the AUC-ROC, as elucidated in \cref{tab:roc}, across various branches. Nevertheless, it is noteworthy that, across all scenarios, AUC-PR consistently registers values lower than AUC-ROC. This can be attributed to the inherent optimism of the AUC-ROC metric in imbalanced data, thereby complicating the translation of AUC-ROC results to real-world scenarios. The observed discrepancy signifies a misclassification event pertaining to the minority class (anomaly instances). Thus, a judicious calibration of the model to achieve an optimal trade-off between precision and recall is imperative, depending upon the specific requirements of the given use case.

% In anomaly detection maintaining a balance between FAR and FRR is often critical to ensure that the system does not miss anomalies (high sensitivity) while also minimizing false alarms (high specificity). EER comes to our aid in finding FAR and FRR when they are in balance. As it can be seen in \cref{tab:all_metrics}, we observe a 34.2\% decrease for the EER on SHT and HR-SHT in pose and combined branches, indicating a more balanced anomaly detection. This reverses on CHAD with the trajectory branch being the most balanced model.

In the field of anomaly detection, achieving a delicate balance between the False Acceptance Rate (FAR) and the False Rejection Rate (FRR) is critical, as it balances the imperative need for high sensitivity to detect anomalies and high specificity to minimize false alarms. The EER serves as the main metric to pinpoint this equilibrium. Notably, as depicted in Table \ref{tab:all_metrics}, a substantial 34.2\% reduction in the EER is observed in both SHT and HR-SHT datasets within the pose and combined models compared to the trajectory, signifying a more harmonious approach to anomaly detection. This trend is reversed in the context of CHAD, where the trajectory branch excels in achieving a balanced anomaly detection model, underscoring the contextual nuances inherent in different environments.

\subsection{Ablation Study}

In this section, we focus on the pose branch of the TSGAD model. As previously discussed, during the training of the GA-VAE, we employ the ELBO loss function. Maximizing ELBO forces the model to acquire more semantically significant representations within the latent space and to achieve a more precise approximation of the true posterior distribution. As a result, it ensures that the reconstructed data closely resembles the original data, preserving faithfulness in reconstruction. 

In the context of unsupervised anomaly detection, when applied to the training dataset consisting of normal videos, a higher ELBO signifies stronger conformity to normal video patterns. After training, when we transition to the test dataset, which contains anomalous frames in addition to normal ones, a low ELBO serves as an indicator of deviations from the learned normal behavior, signifying abnormality. This intrinsic quality of the ELBO can be used as a standalone metric for detecting anomalies.

Therefore, as an alternative to the previously described approach outlined in \cref{sec:methodology}, which involved the construction of a distribution of distributions, we have adopted the utilization of the Evidence Lower Bound (ELBO) inherent to the GA-VAE model as a singular measure. This approach allows us to quantitatively assess the advantages gained through the aforementioned distribution of distributions technique. \cref{tab:ablation} presents a summary of the performance enhancements accomplished through the utilization of the specified technique. In the context of the SHT and HR-SHT datasets, it is evident that GA-VAE outperforms GA-VAE-ELBO in a statistically significant manner. Conversely, in the case of the CHAD dataset, we observe a more subtle differentiation, with GA-VAE-ELBO exhibiting a slight advantage in terms of both AUC-ROC and the AUC-PR, albeit demonstrating inferior performance with respect to the EER.

\section{Conclusion}
\label{sec:conclusion}
Our investigation in this paper delved into the efficacy of variational autoencoders in combination with trajectory prediction for pose-based anomaly detection. Through a series of experiments conducted across multiple benchmark datasets, we have unveiled compelling evidence that this approach holds significant promise. The demonstrated effectiveness of this approach, with consistent performance on diverse datasets, indicates that it represents a worthwhile avenue for future exploration and development within the field of anomaly detection.

\section*{Acknowledgement}
\label{sec:ack}
This research is supported by the National Science Foundation (NSF) under Award Number 2329816.

%%%%%%%%% REFERENCES
{\small
\bibliographystyle{ieee_fullname}
\bibliography{egbib}
}

\end{document}